\documentclass[10pt,journal]{IEEEtran}%
\usepackage{amsfonts}
\usepackage{times}
\usepackage{graphicx}
\usepackage{subfigure}
\usepackage{amsmath}
\usepackage{amssymb}
\usepackage{booktabs}
\usepackage{balance}
\usepackage{hyperref}
\usepackage{caption}
\usepackage{subcaption}
\usepackage{placeins}
\usepackage{algorithm}
\usepackage{algpseudocode}
\usepackage{comment}
\usepackage{multirow}%
\providecommand{\U}[1]{\protect\rule{.1in}{.1in}}
\begin{document}

\title{Relative Age Estimation Using Face Images}
\author{Ran Sandhaus, Yosi Keller$^{\ast}$
\IEEEcompsocitemizethanks{\IEEEcompsocthanksitem R. Sandhaus and Y. Keller are with the Faculty of Engineering, Bar Ilan University, Ramat-Gan, Israel.\protect \and
Email: yosi.keller@gmail.com}}
\maketitle

\begin{abstract}
This work introduces a novel deep-learning approach for estimating age from a
single facial image by refining an initial age estimate. The refinement
leverages a reference face database of individuals with similar ages and
appearances. We employ a network that estimates age differences between an
input image and reference images with known ages, thus refining the initial
estimate. Our method explicitly models age-dependent facial variations using
differential regression, yielding improved accuracy compared to conventional
absolute age estimation. Additionally, we introduce an age augmentation scheme
that iteratively refines initial age estimates by modeling their error
distribution during training. This iterative approach further enhances the
initial estimates. Our approach surpasses existing methods, achieving
state-of-the-art accuracy on the MORPH II and CACD datasets. Furthermore, we
examine the biases inherent in contemporary state-of-the-art age estimation techniques.
\end{abstract}

\section{Introduction}

\label{sec:introduction}

Facial images are a primary modality for age estimation in human perception
and automated systems. In computer vision and biometrics, significant research
has focused on age estimation from facial images, with applications in
e-commerce \cite{Hakeem2012VideoAF}, facial recognition
\cite{Lanitis2004ComparingDC}, and age-based data retrieval. However,
accurately estimating age from facial images remains challenging due to the
complex and heterogeneous facial transformations occurring due to aging.
Ethnicity, gender, and lifestyle are significant factors influencing these
changes. Aging results in progressive facial features and appearance
transformations, where individuals of similar ages often appear comparable,
while those of larger age differences result in more pronounced distinctions
\cite{Facial_Aging}.

To address these challenges, face-based age estimation is typically formulated
as a classification task, assigning a facial query image $\mathbf{x}_{q}$ to
discrete age categories $\{a_{c}\}_{1}^{C}$
\cite{eidinger2014age,guo2011simultaneous,chang2011ordinal,ZhengSun2012,ramon2012gender,guo2010human}%
, or as a regression task, treating age as a continuous variable
$a\in\mathbb{R}^{+}$
\cite{guo2011simultaneous,guo2010human,eidinger2014age,wang2015age,ChenGong2013,leviage}%
. \begin{figure}[tbh]
\begin{center}
\centering\includegraphics[width=0.5\textwidth]{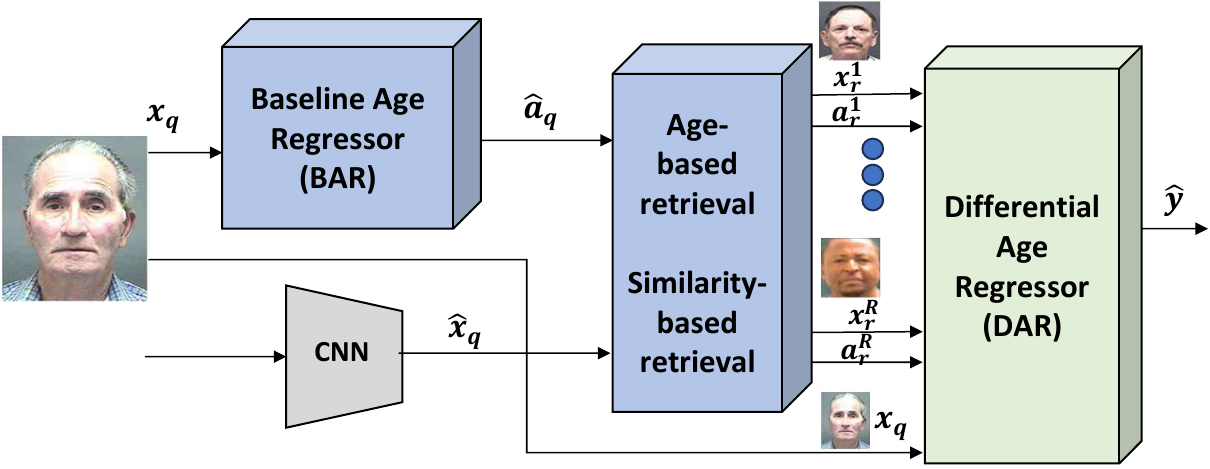}
\end{center}
\caption{\textbf{Differential age estimation.} The Baseline Age Regressor
(BAR) estimates the age $\widehat{a}_{q}$ of the input image $\mathbf{x}_{q}$.
$\widehat{\mathbf{x}}_{q}$ is the CNN embedding of $\mathbf{x}_{q}$ used with
$\widehat{a}_{q}$ to retrieve the set of reference images that are of age
$\widehat{a}_{q}$ and most visually similar to $\mathbf{x}_{q}$. The ages of
the reference images are known. The Differential Age Regressor (DAR) estimates
the age differences between $\mathbf{x}_{q}$ and the reference images and uses
them to refine $\widehat{a}_{q}$.}%
\label{fig:teaser}%
\end{figure}Face-based biometric analysis starts by aligning a facial image to
a canonical spatial frame~\cite{ramon2012gender}, followed by analyzing the
cropped region of interest. Early methods utilized local image descriptors
\cite{ramon2012gender} to encode facial images into high-dimensional
representations for age regression via kernel PLS \cite{guo2011simultaneous}.

In the past decade, advancements in deep learning have enabled the development
of end-to-end trainable age estimation schemes~\cite{8017500, deepage} that
utilize classification and regression losses. Metric learning approaches have
been employed in both shallow~\cite{1640964} and CNN-based
schemes~\cite{8017500,8578147,TianCCY19}, where local features of facial
images were learned by treating age difference as a metric measure. In
contrast, ranking-based
approaches~\cite{8099569,7780901,9145576,Dark_Knowledge} leveraged ordinal
classification to exploit the ordinal structure of age labels for improved
accuracy. Recent methods have integrated classification and regression
techniques~\cite{deepage2}, as well as attention-based
approaches~\cite{deepage2,9673115,lin2021fpage,Ali_2024_WACV}. Despite these
advancements, age estimation has predominantly been approached as predicting a
person's age solely from their facial image. While estimating age differences
between facial images has the potential to improve accuracy by leveraging the
continuous nature of aging, this approach has not yet been integrated into an
absolute age estimation framework.

In this work, we propose a novel approach to age estimation, illustrated in
Fig.~\ref{fig:teaser}. Our method enhances a Baseline Age Regression (BAR) by
training a Differential Age Regression (DAR) model using nearest-neighbor (NN)
reference facial images from the training set. Estimating age differences
using images within small age ranges improves accuracy by addressing the
age-varying characteristics of face-based age estimation. State-of-the-art
(SOTA) BARs typically achieve an accuracy of $\Delta_{BAR}\sim\lbrack15,70]$,
while our DAR model focuses on estimating residual errors within the narrower
range of $\Delta_{DAR}\sim\lbrack-3,+3]$, thereby refining the BAR estimate.
Given the complexity and variability of visual aging patterns in facial
images, and considering that $\Delta_{BAR}\gg\Delta_{DAR}$, our DAR model
demonstrates superior accuracy compared to traditional BAR methods. To our
knowledge, this is the first differential-based age estimation method that
evaluates age differences between a query image and a set of reference images.
For a given query face image $\boldsymbol{\textbf{x}_q}$, we retrieve
reference images $\left\{  \boldsymbol{\mathbf{x}_{r}}\right\}  _{1}^{R}$ from
the training set based on the BAR estimate of $\boldsymbol{\textbf{x}_q}$ and
facial similarity metrics. The DAR model then estimates the age differences
$\left\{  d_{r}\right\}  _{1}^{R}$. The ages of the reference images are known
during both training and testing phases, and the refined age estimate is
obtained as a weighted sum of the estimated age differences. We also propose
an age-augmentation approach, in which, instead of using the BAR directly, we
estimate its error distribution around $a_{q}$, denoted as $D_{\varepsilon}$,
and sample $a_{q}+\varepsilon,$ $\varepsilon\sim D_{\varepsilon}$ to derive an
initial age estimate for retrieving the reference set. Furthermore, we
demonstrate that iteratively estimating $D_{\varepsilon}$ improves the
accuracy of the DAR model. The proposed differential age estimation framework
is model-agnostic and can be integrated with any general regression model to
enhance prediction accuracy.

To conclude, we summarize our contributions as follows:

\begin{itemize}
\item We introduce a novel differential-based age estimation method by
estimating age differences between facial images.

\item By estimating the age differences between an input image and a set of
reference images, we derive a robust and accurate age estimate.

\item We propose an age-augmentation approach that models the error
distribution of the underlying BAR age estimator and samples it to enhance accuracy.

\item The differential age estimation process is iteratively improved by
refining the error distribution over training epochs.

\item The proposed scheme achieves state-of-the-art (SOTA) accuracy on the
MORPH II \cite{1613043} and CACD \cite{10.1007/978-3-319-10599-4_49} age
estimation datasets.
\end{itemize}

\section{Related Work}

Facial age estimation is inherently complex due to significant variations in
aging characteristics across ethnicities, genders, and
lifestyles~\cite{han2013age}. Buolamwini and Gebru~\cite{Gebru} demonstrated
the significance of ethnicity and gender in face analysis and recognition. Guo
and Mu~\cite{guo2010human} introduced a hierarchical approach wherein facial
images are first classified by gender and ethnicity, followed by age
estimation within each subgroup to enhance prediction accuracy. Earlier
approaches relied on local image features to embed facial images, followed by
statistical inference. Balmaseda et al.~\cite{ramon2012gender} used Local
Binary Pattern (LBP) features and SVM classifiers to analyze multiscale
normalized face images and their local context. Zheng and
Sun~\cite{ZhengSun2012} employed a ranking SVM to estimate age by learning
ranking relationships, which were then applied to a reference set for age
estimation. A gender and age classification scheme was introduced by Eidinger
et al.~\cite{eidinger2014age} for non-frontal facial images captured under
uncontrolled conditions. Regression-based approaches reformulate age
estimation as a scalar regression problem using high-dimensional image
embeddings. Thus, a regression model for unbalanced and sparse data was
proposed by Chen and Gong~\cite{ChenGong2013}, enabling accurate age and crowd
density estimation. Low-level visual features extracted from unbalanced and
sparse images were mapped onto a cumulative attribute space, where each
dimension corresponds to a semantic interpretation.

While early methods relied on handcrafted features and statistical models, the
advent of deep learning has significantly transformed facial age estimation,
enabling more robust and data-driven approaches. A hierarchical unsupervised
neural network model was introduced by Wang and Kamikaze~\cite{wang2015age} to
extract robust facial representations. These features were subsequently
processed by Recursive Neural Networks (RNNs) to capture age progression
patterns. Manifold learning was applied to capture the underlying facial aging
manifold by projecting the feature vector into a lower-dimensional, more
discriminative subspace. Hasner and Levi~\cite{leviage} improved the accuracy
of age estimation by formulating it as a classification problem and leveraging
Convolutional Neural Networks (CNNs), while Sendik and Keller~\cite{deepage}
applied deep metric learning to CNN-computed facial features and employed a
Support Vector Regressor (SVR) for age estimation. Deep metric-learning was
also used by Lieu et al.~\cite{8099569} who introduced a hard quadruplet
mining scheme to enhance embeddings, applying a regression-based loss for age
estimation. Rote et al.~\cite{7406390} developed a classification scheme in
which the class probability distribution from the Softmax function was used to
compute the empirical expectancy of the estimated age. Pan et
al.~\cite{Mean-Variance} proposed a multitask approach, where the empirical
probability of each age was computed using the Softmax activation function.
They minimized both the $L_{2}$ loss and the empirical variance of the age
estimation error. A set of CNN-based classification models was suggested by
Malli et al.~\cite{7406402}. Each model was trained to classify within a
specific age range. The final age estimate was obtained by averaging the
outputs of these models.

Shen et al.~\cite{8578343} proposed a hybrid Deep Regression Forests approach
that combines Regression Forests and deep learning inference. In this method,
the forest nodes, which learn adaptive data partitions from the input, are
connected to fully connected layers of a Convolutional Neural Network (CNN).
The Random Forests and CNN are optimized jointly in an end-to-end manner. a
tree-based structure was introduced by Li et al.~\cite{8954134} where adjacent
tree leaves in close branches were connected to create a continuous
transition. Additionally, they employed an ensemble of local regressors, with
each leaf linked to a specific local regressor. The age labels in this
approach were encoded using an ordinal-preserving
representation~\cite{8099569,7780901,9145576,coral} to exploit the inherent
order of age labels. This encoding ensures that each model outputs signals
indicating whether an estimated age exceeds a given threshold. These methods
have been shown to improve the accuracy of age classification.

Niu et al.~\cite{7780901} employed an ordinal regression Convolutional Neural
Network (CNN) to address non-stationarity in aging patterns and develop the
Asian Face Age Dataset (AFAD), which contains more than 160,000 images with
accurately labeled ages. The Deep Cross-Population (DC) domain adaptation
approach by Li et al.~\cite{8578147} for age estimation trains a CNN on a
large source dataset to enhance the accuracy of age estimation on a smaller
target dataset. In the DC approach, transferable aging features are learned
from the source dataset and then transferred to the target dataset.
Additionally, an order-preserving pairwise loss function is utilized to align
the aging features of the two populations. Tain et al.~\cite{TianCCY19}
proposed a correlation learning method to represent and utilize inter- and
intra-cumulative attribute relationships, which was further extended to
perform gender-aware age estimations by leveraging correlations both between
and within gender groups.

Attention-based learning has revolutionized NLP-related tasks and was also
adapted for computer vision. Hiba and Keller \cite{deepage2} introduced a Deep
Learning framework for age estimation, featuring an attention-based image
augmentation-aggregation approach and a hierarchical probabilistic regression
model. While this approach used attention on top of the augmentations, Wang et
al. \cite{9673115} used attention to identify image patches that should be
focused on for age estimation, creating a framework of two CNNs: Attention and
Fusionist. Attention employs a novel OMAHA (Ranking-guided Multi-Head Hybrid
Attention) mechanism to dynamically locate and rank age-specific patches,
which Fusionist integrates with facial images to predict subject age. Line et
al. \cite{lin2021fpage} presented an age estimation method for in-the-wild
scenarios, incorporating facial semantics through a face parsing-based network
and attention module. Considering related tasks in video processing, Deformer,
a video-based model for age classification was proposed by Ali et al.
\cite{Ali_2024_WACV}, that categorizes individuals into four age groups.
Addressing challenges like occlusions and low resolution, the method employs a
two-stream architecture with the Transformer and EfficientNet architectures.

Sun et al. \cite{9541205} addressed age estimation challenges such as
illumination, pose, expression, and the ambiguity of the age labels between
demographic groups. They proposed a general label distribution learning (DLL)
formulation that unifies various age estimation methods. Introducing a deep
conditional distribution learning (DL) method within this framework, the
authors utilized auxiliary face attributes to learn age-related features. From
another perspective, considering the inherent imbalance prevalent across
datasets, Boa et al. \cite{bao2023general} proposed a unified framework for
facial age estimation, addressing challenges in both general and long-tailed
age estimation. They introduced feature rearrangement, pixel-level adjunct
learning, and adaptive routing to enhance performance across diverse age
classes. Siamese graph learning (SGD) was introduced by Lieu et al.
\cite{10068268} to address aging dataset bias. SGD aligns sparse and dense
distributions, preserving the smoothness of aging. The approach employs a
blending strategy for plausible hallucinatory sample generation using
unlabeled data and introduces graph contrastive regularization to mitigate
noise from auxiliary samples. Generative AI as in Delta Age AGAIN (DAA), was
proposed by Chen et al. \cite{chen2023daa} for age recognition using transfer
learning. The DAA operation based on mean and standard deviation values of
style maps employs binary code mapping and a FaceEncoder-AgeDecoder framework.

Several image datasets have been used in face-based age estimation. Some older
datasets, such as FERET \cite{PHILLIPS1998295} (14K images), FG-NET
\cite{cootes2008fg} (1K images), Chalearn LAP 2015
\cite{agustsson2017appareal} (7.5K images), and UTKFace \cite{zhifei2017cvpr}
(16K images), are too small for CNN-based approaches, while others, like
IMDB-Wiki \cite{7406390}, are based on web scraping and human age annotators
without objective groundtruth age labels. As the accuracy of computational age
estimation improves (MAE $\approx$ 2.5 years), it becomes comparable to human
annotations, limiting their effectiveness for future work. The MORPH Album II
\cite{RicanekJr.:2006:MLI:1126250.1126361} is notable as it provides accurate
age and identity labels, while other datasets, including AFAD \cite{7780901},
do not provide identity labels. This has led some works to use only the
Random-Split (RS) test protocol, where the image set is randomly split into
train and test subsets. As most datasets have 25 images of each subject, this
inevitably results in significant train-to-test leakage, making their age
estimation results less reliable. In this work, we focus on datasets equipped
with identity labels, and use the Subject-Exclusive (SE) protocol, where all
of a particular subject's face images are used just in either the train or
test sets.\begin{figure}[tbh]
\begin{center}
\centering\includegraphics[width=0.5\textwidth]{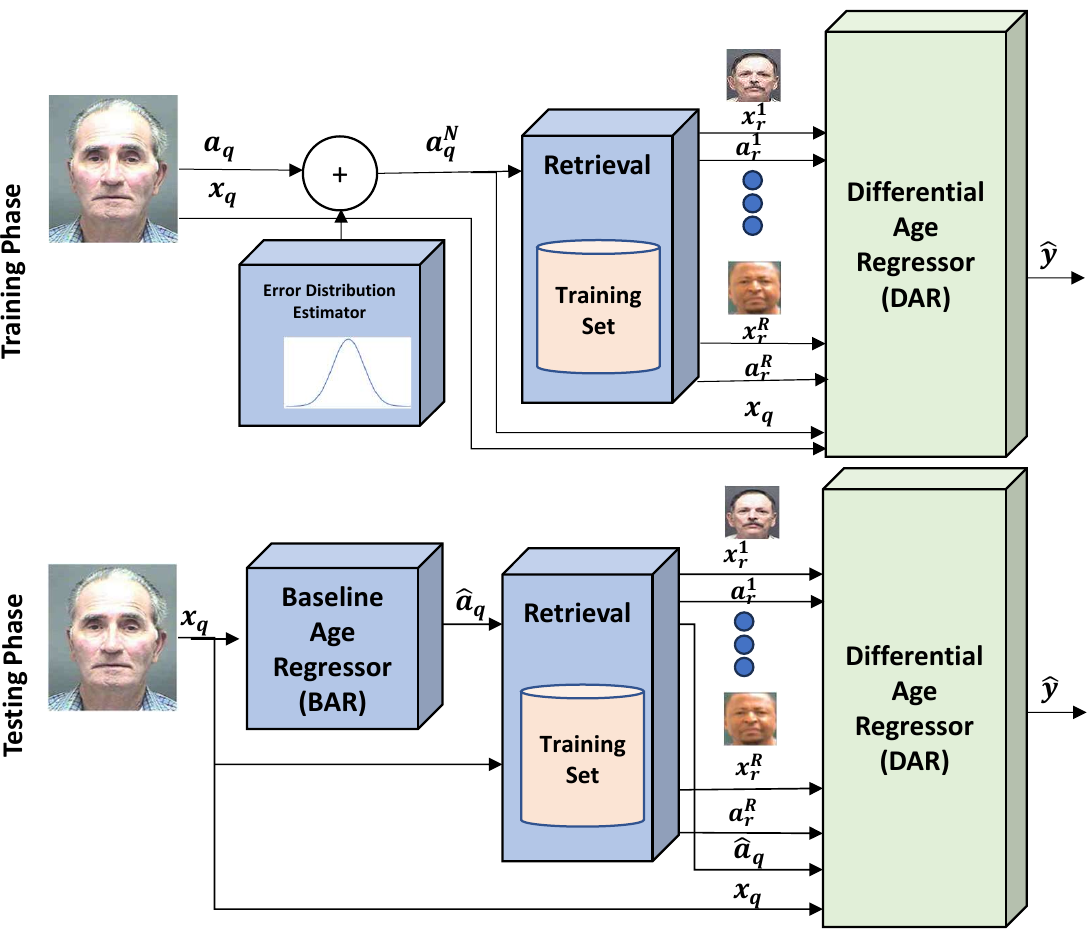}
\end{center}
\caption{\textbf{Differential age estimation in the training and test phases.}
In training time, the groundtruth age $a_{q}$ of the input face image
$\mathbf{x}_{q}$, is augmented using Eq. \ref{equ:train sampling} and used to
retrieve the reference set of images $\{\boldsymbol{x_{r}}\}_{1}^{R}$. In the
test phase, a Baseline Age Regressor (BAR) estimates $\hat{a}_{q}$, the age of
$\mathbf{x}_{q}$ that is used to retrieve $\{\boldsymbol{x_{r}}\}_{1}^{R}$.}%
\label{fig:differential_based_age_est_elaborated}%
\end{figure}\begin{figure*}[tbh]
\begin{center}
\centering\includegraphics[width=0.8\textwidth]{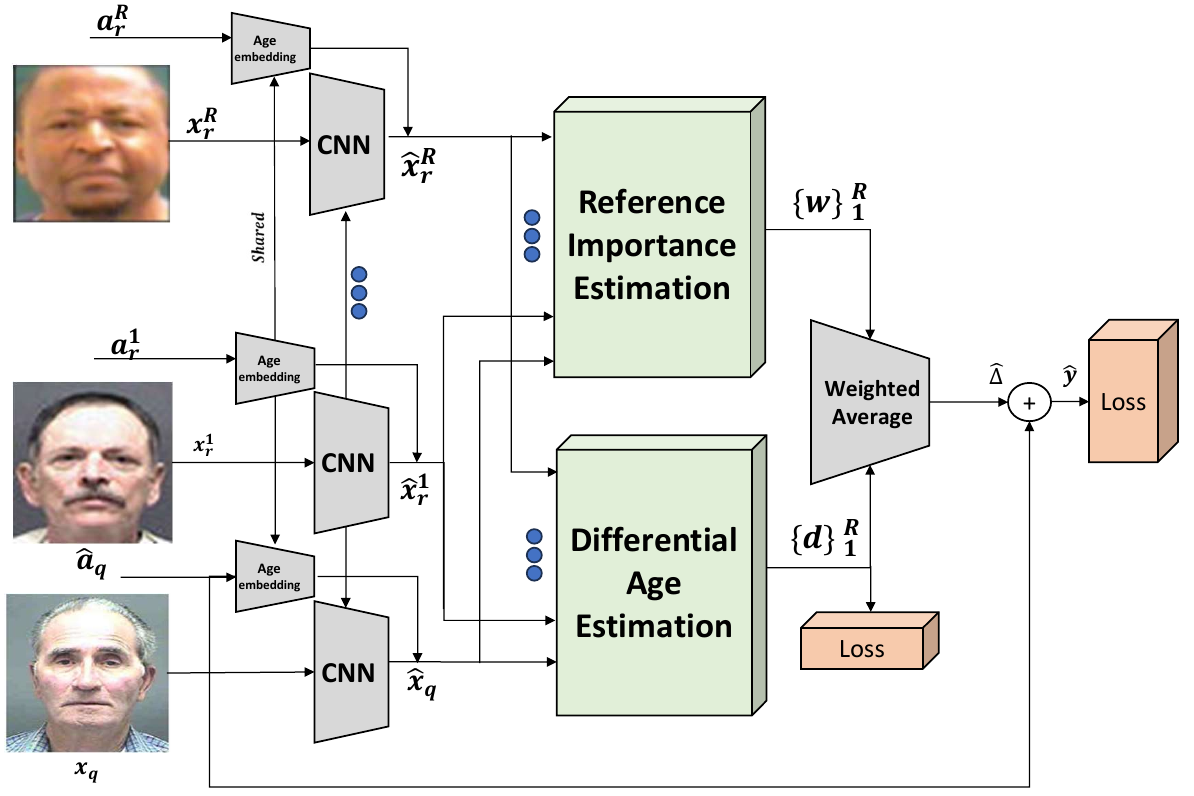}
\end{center}
\par
.\caption{\textbf{The Differential Age Regressor (DAR) network.} The
embeddings of the query image $\boldsymbol{x_{q}}$ and the reference images
$\{\boldsymbol{x_{r}}\}_{1}^{R}$ are computed by a CNN, while the initial age
estimate of the query image $\hat{a}_{q}$ and reference ages
$\{\boldsymbol{a_{r}}\}_{1}^{R}$ are encoded by an embedding layer. The
embeddings are concatenated to $\boldsymbol{{\hat{x}}_{q}}$ and
$\{\boldsymbol{{\hat{x}}_{r}}\}_{1}^{R}$. The DAR network uses the embeddings
to estimate the age differences $\{{d}_{r}\}_{1}^{R}$ and the weights per
reference $\{{w}_{r}\}_{1}^{R}$. The resulting difference estimate
$\hat{\Delta}$ is the weighted average $\hat{\Delta}=\sum_{r}{w}_{r}{d}_{r}$
added to $\hat{a}_{q}$ to compute the age estimate.}%
\label{fig:network_arch_general}%
\end{figure*}

\section{Differential Age Estimation}

\label{sub:dar}

We propose a novel framework for facial age estimation, leveraging
differential age estimation to refine initial predictions. This approach,
illustrated in Fig. \ref{fig:differential_based_age_est_elaborated}, reduces
prediction errors by modeling relative age differences rather than absolute
estimates. Given a query image $\mathbf{x}_{q}$, we first obtain an initial
age estimate $\widehat{a}_{q}$ using a Baseline Age Regressor (BAR). To refine
$\widehat{a}_{q}$, we retrieve a reference set $\{\mathbf{x}_{r}\}_{1}^{R}$ of
individuals with known ages similar to $\mathbf{x}_{q}$. The Differential Age
Regressor (DAR) then estimates the age differences $\{\Delta_{r}\}_{1}^{R}$
between $\mathbf{x}_{q}$ and $\{\mathbf{x}_{r}\}_{1}^{R}$, which are used to
adjust the final prediction. The reference images are retrieved based on two
criteria detailed in Section \ref{subsec:retrieval_desc}: (1) its known age
$a_{r}$ is within a bounded range of $\widehat{a}_{q}$, and (2) it exhibits
high visual similarity to $\mathbf{x}_{q}$ in feature space.

Since initial age estimates are subject to systematic errors, we model the
error distribution $D_{\varepsilon}$ for robust reference selection, and
sample from $a_{q}+\varepsilon,$ $\varepsilon\sim D_{\varepsilon}$. The DAR
estimates the age differences $\{\Delta\mathbf{_{r}}\}_{1}^{R}=\{a_{q}%
-a_{\mathbf{x}_{r}}\}_{1}^{R}$, and the refined age estimate $\widehat{y}$
(Eq. \ref{equ:abs age}) as a weighted sum of $\{\Delta_{r}\}_{1}^{R}$. We use
the BAR by Hiba and Keller~\cite{deepage2} due to its SOTA accuracy. However,
our framework is adaptable and can be integrated with any BAR to refine its
predictions. The DAR model is implemented using the Convolutional Neural
Network (CNN) architecture shown in Fig.~\ref{fig:network_arch_general} and
detailed in Section~\ref{subsec:diff_age_est_detail}. Additionally, in
Section~\ref{subsec:iterative_improvement}, we demonstrate how to iteratively
improve the DAR estimate by refining $D_{\varepsilon}$ and adjusting the
sampling of reference training images.

\subsection{Reference Images Retrieval}

\label{subsec:retrieval_desc}\begin{figure}[tbh]
\begin{center}
\centering\includegraphics[width=0.5\textwidth]{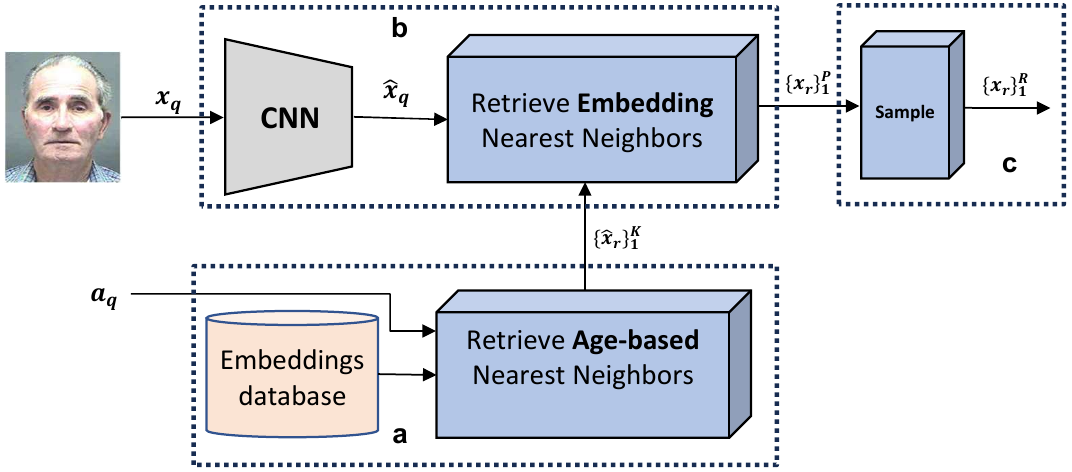}
\end{center}
\caption{\textbf{Reference faces retrieval.} (a) The age $a_{q}$ of
$\boldsymbol{x_{q}}$ is used to retrieve $\{\widehat{\boldsymbol{x}%
}\boldsymbol{_{r}}\}_{1}^{K}$ the embeddings of the reference images of the
same age. (b) $\widehat{\boldsymbol{x}}\boldsymbol{_{q}}$, the embedding of
$\boldsymbol{x_{q}}$ is used to retrieve $\{\widehat{\boldsymbol{x}%
}\boldsymbol{_{r}}\}_{1}^{P},$ $P\ll K,$ the $P$ embeddings in $\{\widehat
{\boldsymbol{x}}\boldsymbol{_{r}}\}_{1}^{K}$ closest to $\widehat
{\boldsymbol{x}}\boldsymbol{_{q}}$. (c) We randomly sample $R<P$ images for
the final reference set $\{\boldsymbol{x_{r}\}}_{1}^{R}$.}%
\label{fig:retrieval}%
\end{figure}

For each query image $\mathbf{x}_{q}$, we retrieve an initial candidate set of
reference images $\{\mathbf{x}_{r}\}_{1}^{K}$ from the training set using the
BAR's estimated age $\widehat{a}_{q}$, as shown in Fig.~\ref{fig:retrieval}.
We then refine this set by selecting the top $P\ll K$ images most visually
similar to $\mathbf{x}_{q}$ using a face embedding network. Directly using
$\widehat{a}_{q}$ for reference retrieval may introduce systematic bias due to
inherent BAR errors. To mitigate this, we estimate an error distribution
$D_{\varepsilon}$ around the true age $a_{q}$, enabling more robust reference
selection:
\begin{equation}
\widehat{a}_{q}=a_{q}+\varepsilon,\quad\varepsilon\sim D_{\varepsilon
}\label{equ:train sampling}%
\end{equation}
and retrieve all reference images $\{\mathbf{x}_{r}\}_{1}^{K}$ whose age is
$\widehat{a}_{q}$. By incorporating $D_{\varepsilon}$, we effectively account
for the BAR's prediction uncertainty, leading to a reference set that is more
representative of possible true ages. Since individuals of the same age can
exhibit significant facial variations due to gender, ethnicity, and lifestyle
factors, we employ a deep face embedding network to refine the reference
selection. This ensures that the final reference set is age-consistent and
visually coherent. Therefore, given the initial age-retrieved reference set
$\{\mathbf{x}_{r}\}_{1}^{K}$, where $K\gg R$, we utilize a face embedding
network \cite{vgg} to retrieve $\{\mathbf{x}_{r}\}_{1}^{P}$,
where $P\gg R$, consisting of the faces most visually similar to
$\mathbf{x}_{q}$. To prevent overfitting and ensure diverse references, we
randomly sample $R$ faces from the visually closest subset $\{\mathbf{x}%
_{r}\}_{1}^{P}$. This balances precision with robustness, improves the
diversity of the training set, and mitigates overfitting. Additionally, we
experimented with alternative sampling methods in
Section~\ref{subsec:ablation}.

The face recognition model is based on the convolutional portion of VGG16,
with the first fully connected (FC) layer removed and a 1D batch normalization
layer added to mitigate overfitting. During testing, the same retrieval
process is repeated without using Eq. ~\ref{equ:train sampling}, ensuring that
$\widehat{a}_{q}=a_{q}$. The BAR's out-of-sample error distribution
$D_{\varepsilon}$ is estimated using Kernel Density Estimation
(KDE)~\cite{silverman1986density}, applied to a subset of 2\% of the dataset
sampled in a randomized, subject-exclusive (SE) manner. In practice, we
restricted $D_{\varepsilon}$ $\in$ $\left[  -20,+20\right]  $, as due to the
limited number of data points where $D_{\varepsilon}$ $\notin\left[
-20,+20\right]  $ the KDE-based estimation was inconsistent.

\subsection{Differential Age Regression Network}

\label{subsec:diff_age_est_detail}

\begin{figure}[tbh]
\includegraphics[width=0.5\textwidth]{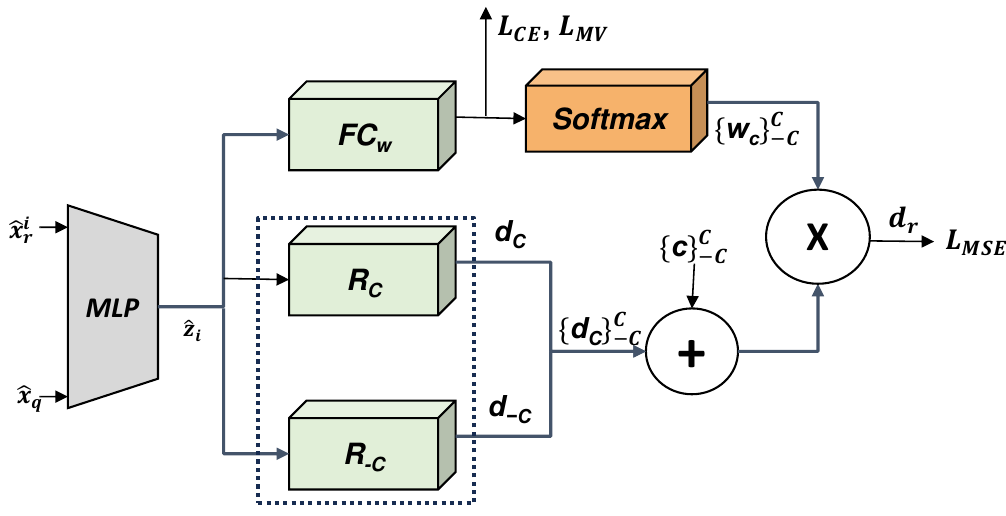}
\caption{\textbf{Differential Age Estimation. }This network estimates the age
differential between the query image $\hat{\mathbf{x}}_{q}$ and a
\textit{single} reference image $\hat{\mathbf{x}}.$ Their embeddings are
concatenated as pairs $\left\{  \hat{\mathbf{x}}_{q},\hat{\mathbf{x}}_{r}%
^{i}\right\}  .$ A set of regressors $\left\{  R_{c}\right\}  _{-C}^{C}$
estimates $\left\{  d_{c}\right\}  _{-C}^{C}:$ the \textit{second-order} age
differentials \textit{around} $\left\{  c\right\}  _{-C}^{C}$. $\left\{
d_{c}\right\}  _{-C}^{C}$ are weighed by $\left\{  w_{c}\right\}  _{-C}^{C}$
computed by $FC_{W}$ and a Softmax layer as in Eq. \ref{equ:diff estimate}.}%
\label{fig:diff_and_importance_subnetworks_arch}%
\end{figure}

The Differential Age Regression (DAR) network estimates age differences
between a query image $\mathbf{x}_{q}$ and a set of reference images
$\{\mathbf{x}_{r}\}_{1}^{R}$. The model is jointly trained to predict the age
differences and the resulting absolute age. The DAR architecture (Fig.
\ref{fig:network_arch_general}) processes input images $\left\{
\mathbf{x}_{q},\{\mathbf{x}_{r}\}_{1}^{R}\right\}  $ through a convolutional
neural network (CNN) backbone. The CNN extracts embeddings, which are
concatenated with age embeddings derived from the reference images' known
ages. This joint representation $\{\hat{\mathbf{x}}_{q},\hat{\mathbf{x}}%
_{r}\}_{1}^{R}$ is passed to the Differential Age Estimation (DAE) network for
age difference prediction. $\{\hat{\mathbf{x}}_{q},\hat{\mathbf{x}}_{r}%
\}_{1}^{R}$ are passed \textit{in parallel} to the Differential Age Estimation
(DAE) network, shown in Fig. \ref{fig:diff_and_importance_subnetworks_arch}.
The DAE estimates the age difference $d_{r}$ and corresponding weight $w_{r}$,
respectively, for \textit{each pair separately}. The final age estimate
$\widehat{y}$ is computed as a weighted sum of predicted age differences
$d_{r}$. The weights $w_{r}$ are learned through the DAE network, prioritizing
references with higher visual similarity and smaller prediction variance:%
\begin{equation}
\widehat{y}=\widehat{a}_{q}+{\sum\limits_{1}^{R}}w_{r}d_{r}%
,\label{equ:abs age}%
\end{equation}
where $\widehat{a}_{q}$ is the BAR's initial age estimate.

In addition to \textit{first-order} age difference estimation $d_{r}$, the
model further refines predictions by computing \textit{second-order
differentials} $d_{c}$, capturing local aging trends. The input images
$\hat{\mathbf{x}}_{q},\hat{\mathbf{x}}_{r}$ to the DAE network (Fig.
\ref{fig:diff_and_importance_subnetworks_arch}) are initially passed through a
3-layer MLP (FC + LeakyReLU + Dropout) with FC layers of dimensions 2048,
1024, and 512, and a Dropout with $p=20\%$. It employs a weighted regression
scheme where a set of regression heads $\left\{  R_{c}\right\}  _{-C}^{C}$,
with $C=20$, estimate the second order age differentials $\left\{
d_{c}\right\}  _{-C}^{C}$ around the difference values $\mathbb{C}%
=\{-C,-C+1,\ldots,C-1,C\}$. $\left\{  d_{c}\right\}  _{-C}^{C}$ are merged by
computing the weights $\left\{  w_{c}\right\}  _{-C}^{C}$ using $FC_{w}$ and a
Softmax such that
\begin{equation}
w_{c}=P\left(  \mathrm{differential}\{\hat{\mathbf{x}}_{q},\hat{\mathbf{x}%
}_{r}\}=c\right)  .\label{equ:DAE class}%
\end{equation}
Thus, the age differential $d_{r}$ between $\hat{\mathbf{x}}_{q}$ and
$\hat{\mathbf{x}}_{r}$ is given by the weighted average:
\begin{equation}
d_{r}={\sum\limits_{-C}^{C}}w_{c}\cdot\left(  c+d_{c}\right)
.\label{equ:diff estimate}%
\end{equation}

Equation \ref{equ:diff estimate} is estimated for all $R$ image pairs
$\{\hat{\mathbf{x}}_{q},\hat{\mathbf{x}}_{r}\}_{1}^{R}$ to compute the
\textit{first-order} differentials $\{d_{r}\}_{1}^{R}$, that are merged in a
weighted average using the weights $\{w_{r}\}_{1}^{R}$ to compute the absolute
age estimate $\widehat{y}$ as in Eq. \ref{equ:abs age}.

\subsection{Model Training}

\label{subsec:training}

Unlike standard absolute regressors, our model simultaneously optimizes
absolute age predictions and relative age differences. This multitask
approach improves generalization by leveraging local and global aging
patterns, leading to more stable and accurate predictions. Thus, the DAR model
is trained using multiple losses. Both the query age estimation $\hat{y}$ and
the age differences $\{d_{r}\}_{1}^{R}$ are jointly optimized, while the
absolute age estimation $\hat{y}$ is optimized with an MSE loss, denoted
$L_{MSE}^{a}$. Each age difference estimate, $d_{r}$, is optimized using
multiple loss terms. The Cross-Entropy Loss $L_{CE}^{d_{i}}$ ensures
classification probabilities in Eq. \ref{equ:DAE class}, while the
Mean-Variance Loss components $L_{M}^{d_{i}}$ and $L_{V}^{d_{i}}$ minimize
prediction variance \cite{Mean-Variance}. The regression results of the
reference age difference are optimized with an MSE loss $L_{MSE}^{d_{i}}$.
Since age differences between two images should be symmetric, we enforce the
constraint $\Delta(\boldsymbol{x_{1}},\boldsymbol{x_{2}})=-\Delta
(\boldsymbol{x_{2}},\boldsymbol{x_{1}})$ This constraint is incorporated
during training by applying the DAR network to both query-reference pairs in
opposite directions, enhancing prediction consistency. After defining
individual loss terms, we combine them into a unified objective function that
balances absolute and differential predictions%
\begin{equation}
L=L_{MSE}^{a}+\frac{1}{R}\sum_{i=1}^{R}(L_{CE}^{d_{i}}+L_{M}^{d_{i}}%
+L_{V}^{d_{i}}+L_{MSE}^{d_{i}}+L_{MSE}^{a}).
\end{equation}
The $L_{MSE}^{d_{i}}$ regression losses of the age difference relate to both
asymmetric age regressions.

\subsection{Iterative DAR Refinement}

\label{subsec:iterative_improvement}

Since the Baseline Age Regressor (BAR) is prone to systematic estimation
errors, we iteratively refine its predictions using the Differential Age
Regressor (DAR). By modeling the BAR's error distribution $D_{\varepsilon}$
and updating it with each iteration, the age estimation process progressively
improves in accuracy. We employ an iterative refinement strategy where the
updated BAR predictions at each step serve as input for the next iteration,
effectively cascading the improvements over multiple refinements
$D_{\varepsilon}^{n}$, progressively reducing prediction errors
\begin{equation}
BAR_{n+1}\left(  D_{\varepsilon}^{n+1}\right)  =DAR\left(  BAR_{n}%
,D_{\varepsilon}^{n}\right)  , \label{equ:iterative dar}%
\end{equation}
where $BAR_{0}$ is the initial absolute age estimator and $D_{\varepsilon}%
^{0}$ as its associated error estimate. At each iteration $n$, the error
distribution $D_{\varepsilon}^{n}$ is updated, refining the BAR predictions
iteratively. This has been experimentally shown in
Section~\ref{subsec:ablation} to improve age estimation accuracy.

\section{Experimental Results}

\label{sec:experimental_results}

The proposed scheme was evaluated using the MORPH II \cite{1613043} and CACD
dataset \cite{10.1007/978-3-319-10599-4_49}, as these large scale datasets
provide the subjects' identities. This allows us to apply the
Subject-Exclusive (SE) protocol. The previously used Random Selection (RS)
evaluation protocol allows images of the same subject to be in both train and
test datasets. This results in train-test leakage, effectively turning the
task into a face-matching problem \cite{deepage2}. The MORPH II \cite{1613043}
is one of the most extensive longitudinal face databases available, containing
55,134 facial images with known identities showing 13,617 subjects between 16
and 77 old. Each subject is shown in multiple mugshots captured under
controlled conditions. The dataset includes individuals of both genders and
diverse ethnicities, primarily White and Black, with demographic and gender
distributions detailed in Table \ref{tab:morph2}. The Cross-Age Celebrity
Dataset (CACD) dataset contains 163,446 images of 2,000 celebrities aged 14 to
62, retrieved from the Internet, with identities provided for each image. The
age was determined by subtracting the celebrity's birth year from the year the
photo was taken. \begin{table}[tbh]
\setlength\tabcolsep{4pt} \centering
\begin{tabular}
[c]{lccccc}%
\toprule Gender & Black & White & Asian & Hispanic & Other\\
\midrule Male & 36,832 & 7,961 & 141 & 1,667 & 44\\
Female & 5,757 & 2,598 & 13 & 102 & 19\\
Total & 42,589 & 10,559 & 154 & 1,769 & 63\\
\bottomrule &  &  &  &  &
\end{tabular}
\caption{Demographic breakdown of the MORPH II \cite{1613043} dataset.}%
\label{tab:morph2}%
\end{table}

We evaluated age estimation accuracy using Mean Absolute Error (MAE),
consistent with prior studies \cite{deepage2}.%
\begin{equation}
MAE=\frac{1}{N}\sum_{i=1}^{N}\lvert\widehat{a}_{q}^{i}-a_{q}^{i}\rvert,
\end{equation}
where $\widehat{a}_{q}^{i}$ and $a_{q}^{i}$ are the predicted and ground truth
age, respectively.

To allow a fair comparison with previous works \cite{deepage2}, the VGG-16
backbone \cite{vgg} was used. Our approach was trained in three
phases. The first, was to fine-tune the backbone (initialized with pre-trained
ImageNet weights) to face recognition using the corresponding training set
(either Morph II \cite{1613043} or CACD \cite{10.1007/978-3-319-10599-4_49})
and the ArcFace loss \cite{Deng_2022}. In the second, we compute the image
embeddings of the VGG-16 backbone. Last, the third phase uses the initial BAR
result to train the proposed DAR approach introduced in Section \ref{sub:dar}.

We used the same training and test sets to train the BAR and DAR models. The
training set is divided into two subject-exclusive parts, as in Section
\ref{subsec:retrieval_desc}: a distribution estimation set (2\% of the
training set) and the model training dataset, used for both the BAR and the
DAR scheme. The input face images are resized to 224$\times$224, intensity
normalized to $[0,255]$, and augmented by randomly applying each with a 0.5
probability: horizontal flipping, color jittering, random affine
transformation, and random small-part erasure. We used $R=10$ references
randomly selected from the nearest neighbor pool of size $P=30$. We used the
Ranger optimizer \cite{Ranger21} and learning rate scheduling using Cosine
Annealing. The experiments were conducted on dual
NVIDIA Tesla V100 GPUs, with the code implemented using the PyTorch framework.
The iterative improvement framework was applied using two iterations of the
training process.

\subsection{Results}

\label{subsec:results}

\begin{figure}[t]
\includegraphics[width=0.5\textwidth]{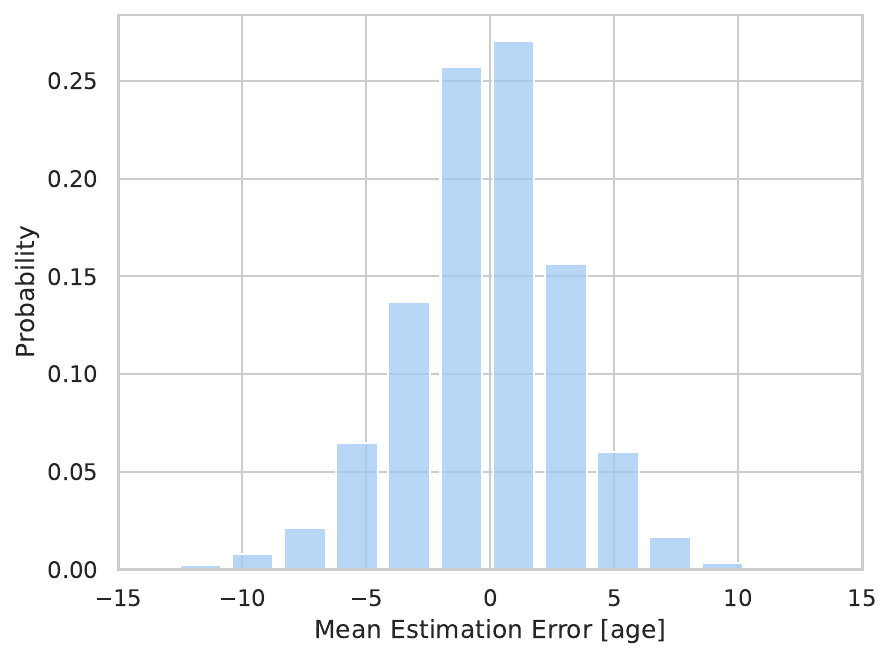}\caption{Distribution
of mean error for the proposed method, over the Morph II dataset.}%
\label{fig:mean_error_hist}%
\end{figure}

\begin{figure}[t]
\includegraphics[width=0.5\textwidth]{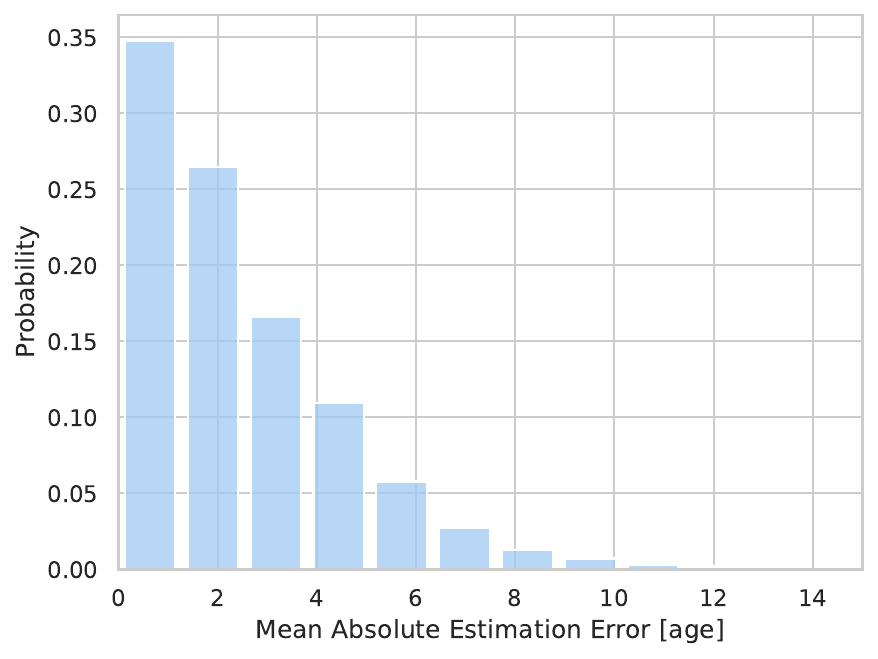}\caption{Distribution
of mean absolute error for the proposed method, over the Morph II dataset.}%
\label{fig:mae_hist}%
\end{figure}

We compared our approach with SOTA methods that are detailed in Tables
\ref{tab:morph2_results} and \ref{tab:cacd_results}. We report their published
results, ensuring adherence to the SE protocol, consistent with our
methodology. We use the same 80\% training and 20\% testing split and SE
protocol as in prior works \cite{deepage2}, where our training set was split
to 78\% training and 2\% of the data for estimating the BAR error distribution
$D_{\varepsilon}$ as in Section \ref{subsec:retrieval_desc}. Table
\ref{tab:morph2_results} presents the MORPH II dataset results, demonstrating
that our approach outperforms previous methods under the SE protocol,
achieving a new state-of-the-art MAE of 2.47. The method also achieves
superior performance on the CACD dataset, with an MAE of 5.27. Figures
\ref{fig:mean_error_hist} and \ref{fig:mae_hist} illustrate the error
distribution of our proposed age estimation scheme, which closely approximates
a Gaussian distribution. \begin{table}[tbh]
\setlength\tabcolsep{4pt} \centering
\begin{tabular}
[c]{llcc}%
\toprule Method & Backbone & MAE & Protocol\\
\midrule Coral\cite{coral} & RESNET34 & 3.27 & SE\\
Mean-Variance\cite{Mean-Variance} & VGG16 & 2.79 & SE\\
soft-ranking\cite{9145576} & RESNET34 & 2.83 & SE\\
soft-ranking\cite{9145576} & VGG16 & 2.71 & SE\\
DCDL\cite{9541205} & VGG-16-BN & 2.62 & SE\\
Hierarchical-Attention\cite{deepage2} & VGG16 & 2.53 & SE\\
Ours & VGG16 & $\mathbf{2.47}$ & SE\\
\bottomrule &  &  &
\end{tabular}
\caption{Age estimation results evaluated using the MORPH II
\cite{1613043} dataset, encompassing our results compared with previous SOTA
approaches using the SE protocol.}%
\label{tab:morph2_results}%
\end{table}\begin{table}[tbh]
\setlength\tabcolsep{4pt} \centering
\begin{tabular}
[c]{lccc}%
\toprule Method & Backbone & MAE & Protocol\\
\midrule Hierarchical-Attention(CNN)\cite{deepage2} & VGG16 & 5.80 & SE\\
Hierarchical-Attention\cite{deepage2} & VGG16 & 5.35 & SE\\
Ours & VGG16 & $\mathbf{5.27}$ & SE\\
\bottomrule &  &  &
\end{tabular}
\caption{Age estimation results evaluated using the CACD
\cite{10.1007/978-3-319-10599-4_49} dataset, encompassing our results compared
with previous SOTA approaches using the SE protocol.}%
\label{tab:cacd_results}%
\end{table}

\subsection{Ablation Study}

\label{subsec:ablation}

We conducted an ablation study to evaluate the contributions of key components
in our approach. In each experiment, a specific algorithmic property or
hyperparameter was systematically varied across a predefined range, followed
by training and evaluation using the MORPH II dataset and the established
protocol. First, we examined the impact of the number of retrieved references
$R$ from the nearest-neighbor pool (size $P=30$), as summarized in Table
\ref{tab:ablation_retrieval_amount_of_refs}. While the number of selected
references influences performance, no clear trend emerges. Interestingly,
increasing the number of references ($R>1$) does not consistently improve
estimation accuracy. \begin{table}[tbh]
\setlength\tabcolsep{4pt} \centering
\begin{tabular}
[c]{cc}%
\toprule \#Refs & MAE\\
\midrule 1 & 2.5\\
$\mathbf{2}$ & $\mathbf{2.47}$\\
5 & 2.51\\
$\mathbf{10}$ & $\mathbf{2.47}$\\
\bottomrule &
\end{tabular}
\caption{Ablation study of the number of references used by the retrieval.}%
\label{tab:ablation_retrieval_amount_of_refs}%
\end{table}

We further analyzed the reference retrieval strategy. After determining the
target reference age, we compared two selection methods: random sampling and
nearest-neighbor-based retrieval. In the random-based method, references are
uniformly selected out of the entire collection of face images of the selected
reference age. In contrast, in the nearest-neighbor-based selection the
references are selected using the $L_{2}$ norm embedding distance (see Section
\ref{subsec:retrieval_desc}). To ensure query-reference dataset diversity and
reduce overfitting, the nearest-neighbor selection first selects a pool of $P$
references, from which a subset is then randomly and uniformly selected. The
results, summarized in Table \ref{tab:ablation_retrieval_method}, indicate
that the nearest-neighbor retrieval method consistently enhances MAE accuracy.
\begin{table}[tbh]
\setlength\tabcolsep{4pt} \centering
\begin{tabular}
[c]{lcc}%
\toprule Retrieval Method & \#Refs & MAE\\
\midrule Random & 10 & 2.5\\
Nearest Neighbor & 10 & $\mathbf{2.47}$\\
\bottomrule &  &
\end{tabular}
\caption{Ablation study of the reference set retrieval approach.}%
\label{tab:ablation_retrieval_method}%
\end{table}

We studied the contribution of the error distribution estimate used to sample
reference training samples in Section \ref{subsec:retrieval_desc}. For that,
we compared two methods. The first is a baseline approach where the references
were sampled from a $\text{U(-3,3)}$ discrete uniform distribution. The second
method uses the proposed KDE-based. The results are summarized in Table
\ref{tab:ablation_error_dist} showing that the KDE approach is superior. We
also found that the uniform distribution-based method converged notably slower
(300 epochs) compared to the KDE-based method (150 epochs).\begin{table}[tbh]
\setlength\tabcolsep{4pt} \centering
\begin{tabular}
[c]{lcc}%
\toprule Error Distribution & \#Refs & MAE\\
\midrule $\text{U(-3,3)}$ & 2 & 2.51\\
KDE & 2 & $\mathbf{2.47}$\\
\bottomrule &  &
\end{tabular}
\caption{Ablation study of the distribution estimate used to sample the age
differences.}%
\label{tab:ablation_error_dist}%
\end{table}

The Differential Age Regression (DAR) network in Section
\ref{subsec:diff_age_est_detail} uses a set of regression heads $\left\{
R_{c}\right\}  _{-C}^{C}$ that estimate the second-order differences and are
probabilistically weighted. In this ablation, we compare this approach to a
simple difference classifier. We show the results of using a single and two
iterations, as in Section \ref{subsec:iterative_improvement}. It follows that
the use of the regression cascade provides a slight, but consistent
improvement.\begin{table}[tbh]
\setlength\tabcolsep{4pt} \centering%
\begin{tabular}
[c]{lcc}%
\toprule\multirow{2}{*}{ADR Method} & \multicolumn{2}{c}{Iteration}\\
& \#1 & \#2\\
\midrule Classifier & 2.49 & 2.48\\
Regressors + Classifier & 2.50 & \textbf{2.4}7\\
\bottomrule &  &
\end{tabular}
\caption{Ablation study of the Differential Age Regression (DAR) network
(Section \ref{subsec:diff_age_est_detail}). We compare our scheme to using an
age-differential classifier, without the additional set of regressors.}%
\label{tab:first_iteration}%
\end{table}

Finally, we evaluated the iterative improvement proposed in Section
\ref{subsec:iterative_improvement}. We present the results of an iterative
improvement using two training refinement steps: training, saving the results
over the distribution set, and retraining based on these results to refine the
first phase. We used KDE and $R=10$ references, as this configuration achieved
SOTA results. The results are summarized in Table
\ref{tab:ablation_iterative_improvement}, and it is worth noting that we also
experimented with multiple learning rates in the second
iteration.\begin{table}[tbh]
\setlength\tabcolsep{4pt} \centering
\begin{tabular}
[c]{lcccc}%
\toprule Error Distribution & \#Refs & Iteration & Learning Rate & MAE\\
\midrule KDE & 10 & 1 & 0.0003 & 2.5\\
$\mathbf{KDE}$ & $\mathbf{10}$ & $\mathbf{2}$ & $\mathbf{0.0003}$ &
$\mathbf{2.47}$\\
\bottomrule &  &  &  &
\end{tabular}
\caption{Ablation study of the iterative refinement using single and two
iterations.}%
\label{tab:ablation_iterative_improvement}%
\end{table}

\section{Bias Analysis}

We conducted a statistical bias analysis of our proposed scheme using the
MORPH II dataset, whose gender and ethnicity distributions are detailed in
Table \ref{tab:morph2}. Our methodology follows the approach outlined by Hiba
and Keller \cite{deepage2}. As the MORPH II dataset exhibits imbalances in
age, gender, and ethnicity, and our approach relies on random sampling, the
resulting training and test sets inherit these biases. We analyze the bias
introduced by our scheme, despite achieving SOTA accuracy under the SE
protocol, to better understand its implications.

\textbf{Age bias.} Table \ref{tab:bias_analysis_age_bias} reports the error
distribution for different age ranges. We report the number of training
samples per age range, as the error relates to the number of training samples.
The lowest error is observed for the 15-24 age range. As the number of
training samples in these bins (15-19 and 20-24) is comparable to some older
age ranges that exhibit higher estimation errors, this suggests that lower
errors are primarily due to appearance variations rather than sample size
alone. However, the number of samples also contributes to the predictive
power. Age estimation error remains relatively stable for midlife ages (30-50)
but increases significantly for older age groups (55+), where the availability
of training samples is substantially lower.

\begin{table}[tbh]
\setlength\tabcolsep{4pt} \centering
\begin{tabular}
[c]{cccc}%
\toprule Age & \#Samples & MAE & STD\\
\midrule 15-19 & 5961 & 1.62 & 1.88\\
20-24 & 7320 & 2.01 & 2.59\\
25-29 & 5619 & 2.49 & 3.15\\
30-34 & 5630 & 2.73 & 3.45\\
35-39 & 6905 & 2.59 & 3.34\\
40-44 & 5772 & 2.64 & 3.29\\
45-49 & 3879 & 3.02 & 3.49\\
50-54 & 2182 & 3.25 & 3.47\\
55-59 & 753 & 4.58 & 3.43\\
60-64 & 192 & 5.67 & 3.09\\
65-69 & 56 & 9.56 & 1.45\\
\bottomrule &  &  &
\end{tabular}
\caption{\textbf{Age bias.} Each row presents a separate age range bin, its
amount of training samples, and the resulting test MAE and standard deviation
of the age estimation error.}%
\label{tab:bias_analysis_age_bias}%
\end{table}

\begin{figure}[th]
\includegraphics[width=0.4\textwidth]{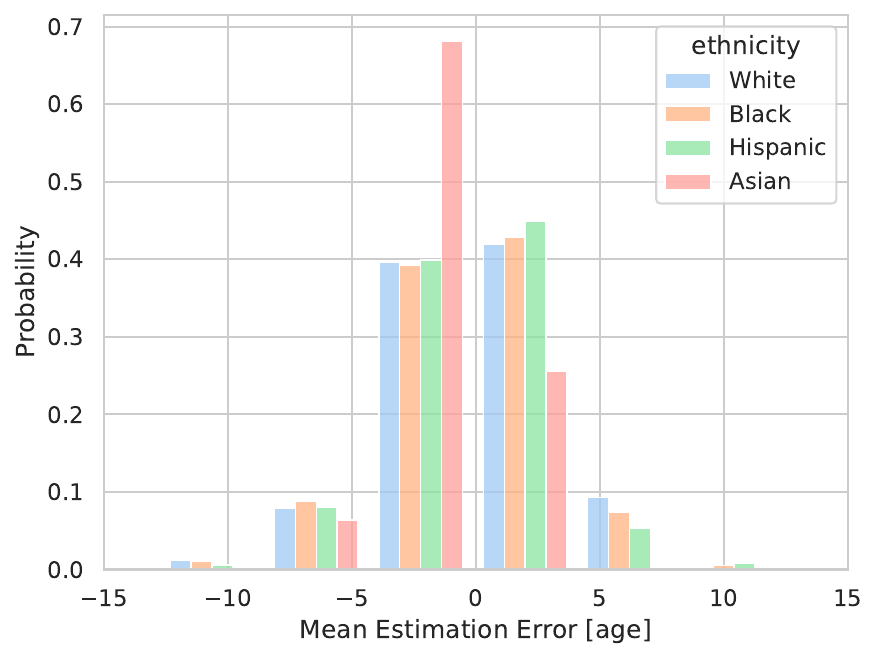}\caption{\textbf{Ethnicity
bias (error histogram)}. Distribution of mean error for our proposed method,
over the Morph II dataset, across the different ethnicities in the dataset.}%
\label{fig:mean_error_hist_per_ethnicity}%
\end{figure}

\begin{figure}[t]
\includegraphics[width=0.4\textwidth, height=150pt]{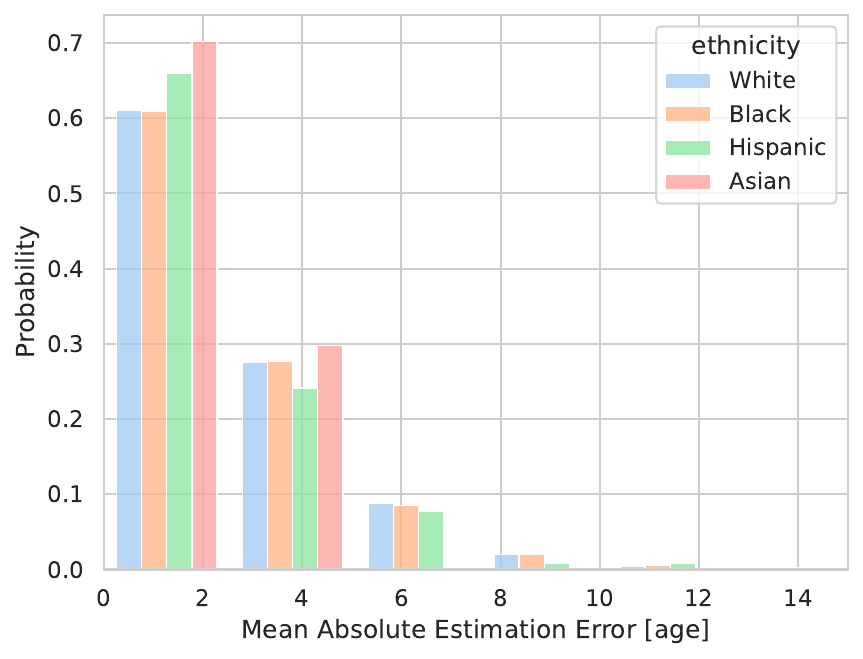}\caption{\textbf{Ethnicity
bias (MAE histogram)}. Distribution of the MAE of our proposed method, using the
Morph II dataset, across the different ethnicities in the dataset.}%
\label{fig:mae_hist_per_ethnicity}%
\end{figure}

\begin{figure}[t]
\includegraphics[width=0.4\textwidth, height=150pt]{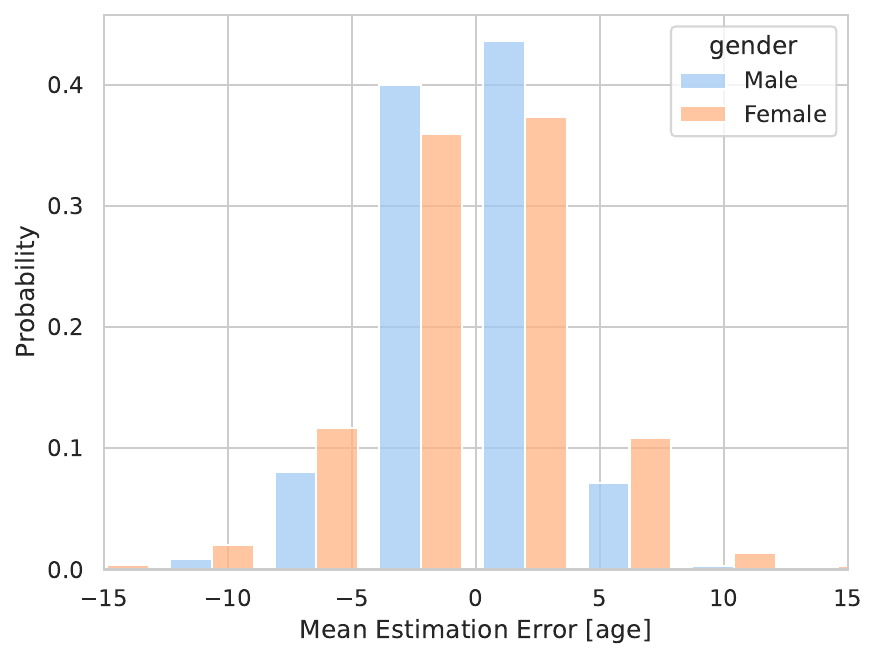}\caption{\textbf{Gender
bias (error histogram)}. A gender-wise distribution of mean error for our
proposed method, using the Morph II dataset.}%
\label{fig:mean_error_hist_per_gender}%
\end{figure}

\begin{figure}[t]
\includegraphics[width=0.4\textwidth, height=150pt]{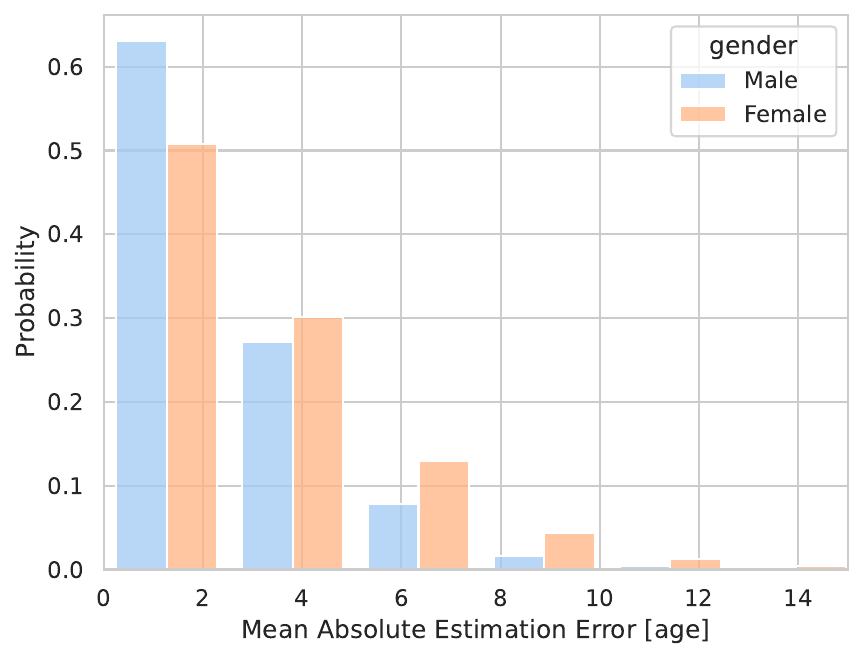}\caption{\textbf{Gender
bias (MAE histogram)}. A gender-wise distribution of MAE for our proposed
method, over the Morph II dataset.}%
\label{fig:mae_hist_per_gender}%
\end{figure}

\textbf{Gender and ethnicity bias.} Gender and ethnicity are the most common
sources of estimation bias in biometrics use cases \cite{robinson2020face}.
Figures \ref{fig:mean_error_hist_per_ethnicity} and
\ref{fig:mae_hist_per_ethnicity} examine the ethnicity bias. We present the
mean error and MAE histogram across all ethnicities. The MORPH II database is
heavily skewed towards Black men, who make up 67\% of the dataset. In Figures
\ref{fig:mean_error_hist_per_gender} and \ref{fig:mae_hist_per_gender}, we
present the gender bias analysis. The MAE for men is about 24\% lower than
that of women, resulting from the larger number of male training samples.


\begin{figure}[tbh]
\includegraphics[width=0.4\textwidth]{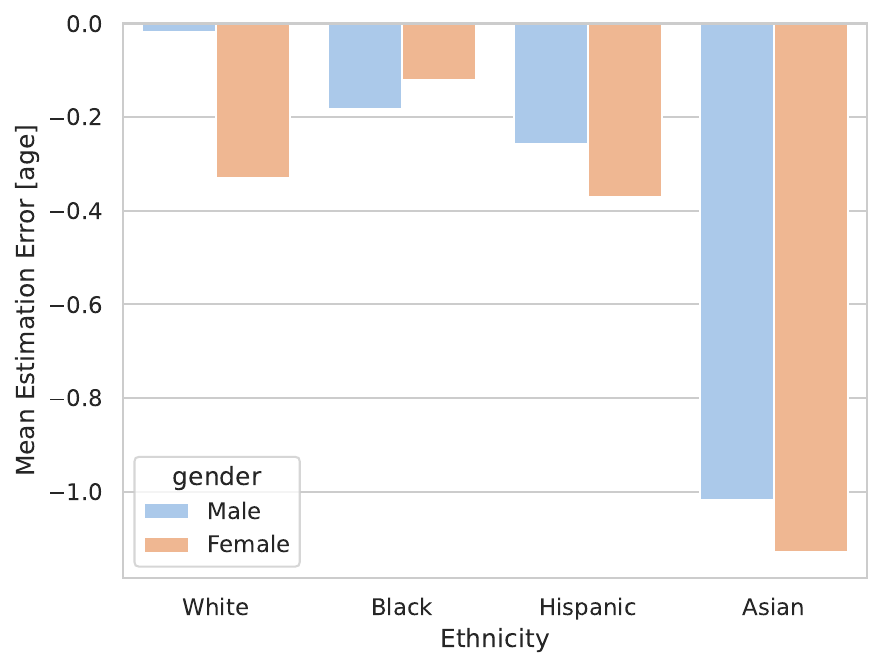}\caption{\textbf{Gender
and ethnicity bias}. A per-ethnicity and gender breakdown of the mean error for
our proposed method, using the Morph II dataset.}%
\label{fig:mean_error_ethnicity_per_gender}%
\end{figure}

\begin{figure}[tbh]
\includegraphics[width=0.4\textwidth]{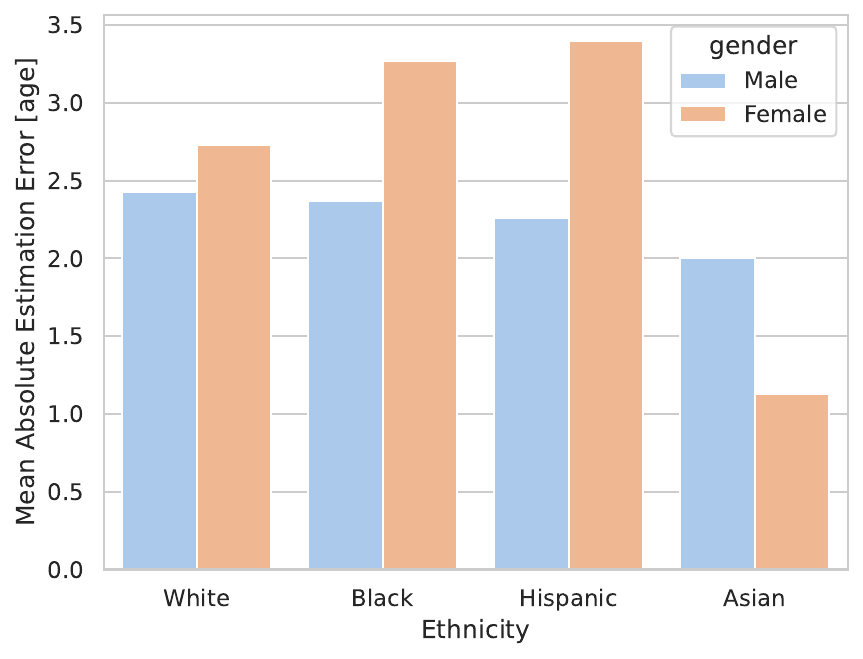}\caption{\textbf{Gender
bias (MAE histogram)}. A per-ethnicity and gender breakdown of MAE for our
proposed method, using the Morph II dataset.}%
\label{fig:mae_per_ethnicity_per_gender}%
\end{figure}

Figures \ref{fig:mean_error_ethnicity_per_gender} and
\ref{fig:mae_per_ethnicity_per_gender} present the estimation bias across both
gender and ethnicity, indicating the error and MAE per each ethnicity class
and gender. The MAE for female subjects is greater across all ethnicities,
except for the Asian ethnicity. The MAE across the various ethnicity classes
is much more uniform, with noticeably smaller variability for men, compared to
women.

\section{Conclusions}

We propose a novel framework for age estimation from facial images. First, we
introduce a differential age estimation approach that trains an age difference
estimator, using a query image and a set of reference images retrieved by a
baseline age estimator. Second, we enhance the baseline age estimator by
Kernel Density Estimation (KDE) to effectively sample reference images,
improving the diversity and relevance of the reference set. The individual age
estimations for each reference image are aggregated using learned
probabilistic weights to produce the final age estimate. To our knowledge,
ours is the first work to present a differential age estimation scheme. We
also propose an iterative refinement of the BAR\ error estimate, which further
enhances the accuracy of the age predictions. Experimental results demonstrate
that our approach outperforms existing SOTA methods.

\balance
\FloatBarrier
{\small
\bibliographystyle{IEEEtran}
\bibliography{diffage}
}

\end{document}